\pdfoutput=1
\documentclass[11pt]{article}

\usepackage{acl}

\usepackage{times}
\usepackage{latexsym}

\usepackage[T1]{fontenc}
\usepackage[utf8]{inputenc}

\usepackage{microtype}

\usepackage{times}
\usepackage{soul}
\usepackage{url}
\usepackage[utf8]{inputenc}
\usepackage{graphicx}
\usepackage{amsmath}
\usepackage{amsthm}
\usepackage{booktabs}
\usepackage{algorithm}
\usepackage{algorithmic}
\usepackage{amsthm,amssymb,mathrsfs}
\usepackage{multirow}
\usepackage{color}
\usepackage{stfloats}
\usepackage{bm}
\usepackage{changes}
\usepackage{makecell}
\usepackage{balance}
\title{Multi-modal Contrastive Representation Learning for Entity Alignment}

\author{
Zhenxi Lin\textsuperscript{\rm 1},
Ziheng Zhang\textsuperscript{\rm 1},
Meng Wang\textsuperscript{\rm 2},
Yinghui Shi\textsuperscript{\rm 3},
Xian Wu\textsuperscript{\rm 1},
Yefeng Zheng\textsuperscript{\rm 1} \\
\textsuperscript{\rm 1}Tencent Jarvis Lab, Shenzhen, China \\
\textsuperscript{\rm 2}School of Computer Science and Engineering, Southeast University, Nanjing, China\\
\textsuperscript{\rm 3}School of Cyber Science and Engineering, Southeast University, Nanjing, China \\
\normalsize{\texttt{\{chalerislin,zihengzhang\}@tencent.com}} \\
\normalsize{\texttt{\{meng.wang,shiyinghui\}@seu.edu.cn}} \\
\normalsize{\texttt{\{kevinxwu,yefengzheng\}@tencent.com}}
}
\begin{document}
\maketitle

\begin{abstract}

Multi-modal entity alignment aims to identify equivalent entities between two different multi-modal knowledge graphs, which consist of structural triples and images associated with entities.
Most previous works focus on how to utilize and encode information from different modalities, while it is not trivial to leverage multi-modal knowledge in entity alignment because of the modality heterogeneity.
In this paper, we propose MCLEA, a \textbf{M}ulti-modal \textbf{C}ontrastive \textbf{L}earning based \textbf{E}ntity \textbf{A}lignment model, to obtain effective joint representations for multi-modal entity alignment.
Different from previous works, MCLEA considers task-oriented modality and models the inter-modal relationships for each entity representation. In particular, MCLEA firstly learns multiple individual representations from multiple modalities, and then performs contrastive learning to jointly model intra-modal and inter-modal interactions.
Extensive experimental results show that MCLEA outperforms state-of-the-art baselines on public datasets under both supervised and unsupervised settings.\footnote{The source code is available at \url{https://github.com/lzxlin/MCLEA}.}

\end{abstract}

\section{Introduction}

Knowledge graphs (KGs) such as DBpedia~\cite{lehmann2015dbpedia} and YAGO~\cite{mahdisoltani2014yago3} employ the graph structure to represent real-world knowledge, where the concepts are represented as nodes and the relationships among concepts are represented as edges.
KGs have been widely applied to knowledge-driven applications to boost their performance, like 
recommendation system~\cite{cao2019unifying}, information extraction~\cite{han2018neural} and question answering~\cite{ijcai2021p611}.
% question answering~\cite{zhang2019multi} and link prediction~\cite{wang2021visual}.
In recent years, an increasing amount of knowledge has been represented in multi-modal formats, such as MMKG~\cite{liu2019mmkg} and Richpedia~\cite{wang2020richpedia}. These multi-modal KGs usually contain images as the visual modality, like profile photos, thumbnails, or posters. The augmented visual modality has shown the significant capability to improve KG-based applications~\cite{chen2020mmea}.
It was also proven that the incorporation of visual modality can enhance the contextual semantics of entities and also achieve improved KG embeddings~\cite{wang2021visual}.

Due to the large scope of real-world knowledge, most KGs are often incomplete, and multiple different KGs are usually complementary to one another.
As a result, integrating multiple KGs into a unified one can enlarge the knowledge coverage and can also assist in refining KG by discovering the potential flaws~\cite{chen2020mmea}.
To integrate heterogeneous multi-modal knowledge graphs, the task of multi-modal entity alignment (MMEA) is therefore proposed, which aims to discover equivalent entities referring to the same real-world object.
Several previous MMEA works have shown that the inclusion of visual modality in modeling helps to improve the performance of entity alignment. For instance, MMEA~\cite{chen2020mmea} and EVA~\cite{liu2021visual} proposed distinct multi-modal fusion modules to integrate entity representations from multiple modalities into joint embeddings and they achieved state-of-the-art performance.\footnote{To distinguish the model MMEA from the task MMEA, we use EA to denote multi-modal entity alignment for the rest of the paper.}
However, these methods mainly utilize existing representations from different modalities, and customized representation learning for EA is not fully explored. In addition, existing methods only explore the use of diverse multi-modal representations to enhance the contextual embedding of entities, the inter-modal interactions are often neglected in modeling.

To address aforementioned problems, we propose MCLEA, a \textbf{M}ulti-modal \textbf{C}ontrastive \textbf{L}earning based \textbf{E}ntity \textbf{A}lignment model, which effectively integrates multi-modal information into joint representations for EA.
MCLEA firstly utilizes multiple individual encoders to obtain modality-specific representations for each entity. The individually encoded information includes neighborhood structures, relations, attributes, surface forms (i.e., entity names), and images.
Furthermore, we introduce contrastive learning into EA with intra-modal contrastive loss and inter-modal alignment loss.
Specifically, intra-modal contrastive loss aims at distinguishing the embeddings of equivalent entities from the ones of other entities for each modality, thus generating more appropriate representations for EA.
Inter-modal alignment loss, on the other hand, aims at modelling inter-modal interactions and reducing the gaps between modalities for each entity.
With these two losses, MCLEA can learn discriminative cross-modal latent embeddings and ensure potentially equivalent entities close in the joint embedding space, regardless of the modality.
MCLEA is also generic as it can support a wide variety of modalities.
Moreover, it combines multiple losses and simultaneously learns multiple objectives using task-dependent uncertainty.

The contributions of this paper are three-fold: 
% (\textit{i}) We propose MCLEA that embeds information from different modalities into a unified vector space and then obtains discriminative entity representations based on contrastive learning for entity alignment.
(\textit{i}) We propose a novel method, called MCLEA, to embed information from different modalities into a unified vector space and then obtain discriminative entity representations based on contrastive learning for entity alignment.
(\textit{ii}) We propose two novel losses to explore intra-modal relationships and inter-modal interactions, ensuring that to-be-aligned entities between different KGs are semantically close with minimum gaps between modalities.
(\textit{iii}) We experimentally validate the effectiveness and superiority of MCLEA as it achieves state-of-the-art performance on several public datasets in both supervised and unsupervised settings. The overall results also suggest that our MCLEA is capable of learning more discriminative embedding space for multi-modal entity alignment.

\section{Related Work}

\subsection{Multi-modal Knowledge Graph}

While many efforts~\cite{mahdisoltani2014yago3,lehmann2015dbpedia} have been made to achieve large-scale KGs, there are just a few attempts to enrich KGs with multiple modalities.
For example, MMKG~\cite{liu2019mmkg} and Richpedia~\cite{wang2020richpedia} utilized the rich visual resources (mainly images) to construct multi-modal knowledge graphs.
Their target was to enrich the KG information via appending sufficient and diverse images to textual entities but they also brought challenges to the KG embedding methods.
Unlike traditional KG embedding methods, multi-modal KG embedding methods model the textual and visual modalities at the same time~\cite{zhang2019multi,wang2021visual}.
For example, \citet{zhang2019multi} proposed MKHAN to exploit multi-modal KGs with hierarchical attention networks on question answering, and \citet{wang2021visual} proposed RSME to selectively incorporate visual information during the KG embedding learning process.
% \todo{There are many more attempts on multi-modal KG embedding, but they all share the single-KG assumption and they are compatible with learning the embeddings of two KGs and also discovering the equivalent cross-KG mappings.}

\begin{figure*}[!h]
    \centering
    \includegraphics[width=1.0\linewidth]{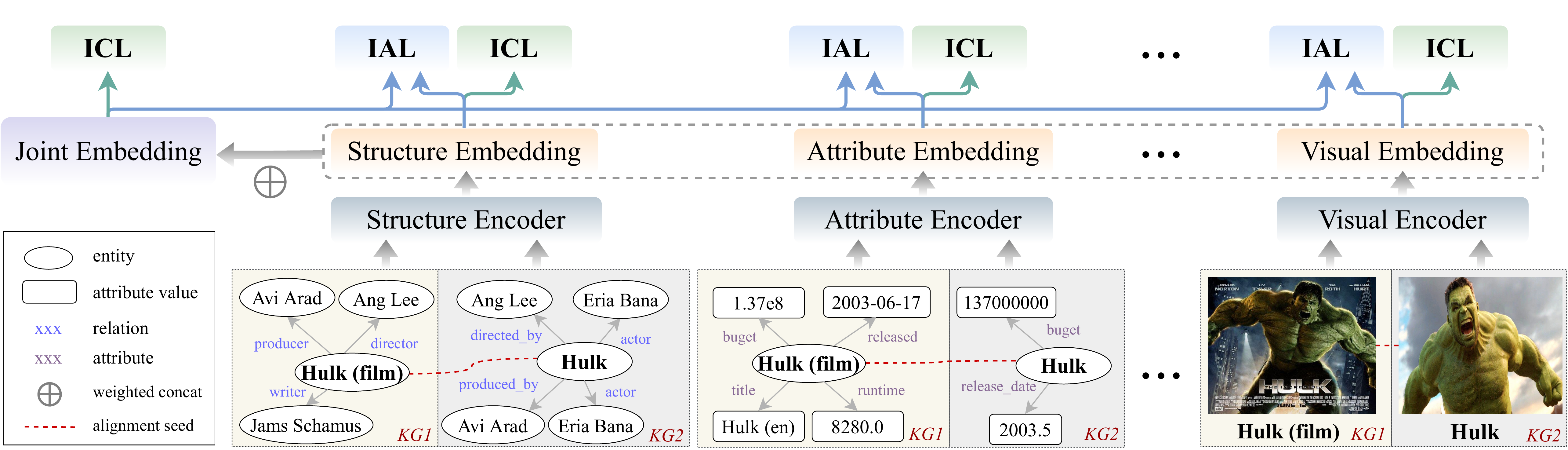}
    \caption{The overall architecture of MCLEA, which combines multiple modalities (\S~\ref{sec:embedding}) and learns through two proposed losses (\S~\ref{sec:cl}), intra-modal contrastive loss (ICL) and inter-modal alignment loss (IAL).}
    \label{fig:model}
\end{figure*}

\subsection{Multi-modal Entity Alignment}

Recent studies for entity alignment mostly focused on exploring the symbolic similarities based on various features, including entity names~\cite{wu2019relation,zhang2019multi}, 
% entity types~\cite{xiang2021ontoea}, 
attributes~\cite{liu2020exploring}, descriptions~\cite{zhang2019multi,tang2021bert} and ontologies~\cite{xiang2021ontoea}.
Most of them started with transforming entities from different KGs into a unified low-dimensional vector space by translation-based models or graph neural networks and then discovered their counterparts based on the similarity metrics of entity embeddings.
Some surveys summarized that additional KG information, if appropriately encoded, could further improve the performance of EA methods~\cite{ZQW2020VLDB,ZJX2020COLING}.
Some previous attempts even proposed to guide these embedding-based EA models with probabilistic reasoning~\cite{qi2021unsupervised}.
With such findings and the increasing popularity of multi-modal knowledge graphs, how to incorporate visual modality in EA, namely multi-modal entity alignment, has begun to draw research attention but the attempts are limited.
The pioneer method PoE~\cite{liu2019mmkg} combined the multi-modal features and measured the credibility of facts by matching the underlying semantics of entities.
Afterward, MMEA~\cite{chen2020mmea} was proposed to integrate knowledge from different modalities (relational, visual, and numerical) and obtain the joint entity representations.
Another method termed EVA~\cite{liu2021visual} leveraged visual knowledge and other auxiliary information to achieve EA in both supervised and unsupervised manner.
% It is worth noting that the cross-modal interaction was achieved with fusion modules in both MMEA and EVA, which certainly cannot obtain task-oriented representations.
Alternatively, the method HMEA~\cite{guo2021multi} modeled structural and visual representations in the hyperbolic space, while Masked-MMEA~\cite{shi2022prob} discussed the impacts of visual modality and proposed to incorporate a selectively masking technique to filter potential visual noises.
These methods mainly utilize multi-modal representations to enhance the contextual embedding of entities, nevertheless, customized entity representations for EA and inter-modal interactions are often neglected in modeling.
Different from previous methods, our proposed MCLEA can learn both intra-modal and inter-modal dynamics simultaneously by the proposed contrastive objectives, expecting to learn more discriminative and abundant entity representation for EA.
% Moreover, MCLEA regards different features as different modalities, and therefore, MCLEA is scalable to support any number of modalities.

\section{Proposed Method}

% In this section, we first explain the notations and the formulation of the problem statement, after which we introduce the multi-modal embeddings and the designed two loss functions based on contrastive learning. Finally, the optimization and training strategy will be explained. We present the overall architecture of MCLEA in Figure~\ref{fig:model}.

% This section starts with the problem formulation and the notations in \S~\ref{sec:problem}. The overall architecture of the proposed MCLEA is shown in Figure~\ref{fig:model}, and its primary components, multi-modal embeddings and contrastive representation learning, are detailed in \S~\ref{sec:embedding} and \S~\ref{sec:cl}, respectively. The optimization and training strategy is explained in \S~\ref{sec:objective}.

We start with the problem formulation and the notations.
A multi-modal KG is denoted as $G=(E,R,A,V,T)$, where $E,R,A,V,T$ are the sets of entities, relations, attributes, images, and triples, respectively. 
Given $G_1=(E_1,R_1,A_1,V_1,T_1)$ and $G_2=(E_2,R_2,A_2,V_2,T_2)$ as two KGs to be aligned, the aim of EA is to find aligned entity pairs $\mathcal{A}=\{(e_1,e_2)|e_1\equiv e_2,e_1\in {E_1}, e_2\in {E_2}\}$, where we assume a small set of entity pairs $\mathcal{S}$ (seeds) are given as training data.
The overall architecture of the proposed MCLEA is shown in Figure~\ref{fig:model}, and its primary components, multi-modal embeddings, and contrastive representation learning will be detailed in the following sections.
% , are detailed in \S~\ref{sec:embedding} and \S~\ref{sec:cl}, respectively.
% The optimization and training strategy is explained in \S~\ref{sec:objective}.

\subsection{Multi-Modal Embeddings}
\label{sec:embedding}

Multi-modal KGs often depict various features with multiple modalities (or views), which are complementary to each other. 
% Employing multi-modal features helps better understand the content of entities.
% In this section, we learn a joint entity embedding from each particular modality, where each modality is processed by an encoder network adapted to the nature of the signal.
We investigate different embeddings from different modalities for MCLEA, including neighborhood structures, relations, attributes, names (often termed as surface forms in previous work~\cite{liu2021visual}), and images.
Each modality is processed with an individual encoder network adapted to the nature of the signal. 
Furthermore, these uni-modal embeddings are aggregated with a simple weighted mechanism to form a joint embedding.
% Note that, as aforementioned, MCLEA is capable of supporting any number of modalities.
% Note that MCLEA is scalable to more modalities of KG.
% Note that, as aforementioned, MCLEA is capable of supporting any number of modalities.
Theoretically, MCLEA can support more modalities, e.g., numerical values~\cite{chen2020mmea}, which will be left in our future work.

\subsubsection{Neighborhood Structure Embedding}

The graph attention network (GAT) is a typical neural network that directly deals with structured data~\cite{velivckovic2018graph}.
% Hence, we leverage GAT as the aggregation mechanism to aggregate the neighbors representation of entities in $G_1$ and $G_2$.
Hence, we leverage GAT to model the structural information of $G_1$ and $G_2$, shown as ``\textbf{Structure Encoder}'' in Figure~\ref{fig:model}.
Specifically, the hidden state $\bm{h}_i\in\mathbb{R}^d$ ($d$ is the hidden size) of entity $e_i$ by aggregating its one-hop neighbors $\mathcal{N}_i$ with self-loop is formulated as:
\begin{equation}
    \bm{h}_i = 
    \textstyle\sigma\left(\sum_{j\in{\mathcal{N}_i}}\alpha_{ij}\bm{h}_j\right),
    \label{gat}
\end{equation}
where $\bm{h}_j$ is the hidden state of entity $e_j$, $\sigma(\cdot)$ denotes the ReLU nonlinearity, and $\alpha_{ij}$ denotes the importance of entity $e_j$ to entity $e_i$, which is calculated with the self-attention:
% \begin{small}
\begin{equation}
    \alpha_{ij} = 
    \frac{
    \exp\left(\eta\left(\mathbf{a}^T[\mathbf{W} \bm{h}_i \oplus \mathbf{W} \bm{h}_j]\right)\right)
    }{
    \sum_{u\in\mathcal{N}_i}\exp\left(\eta\left(\mathbf{a}^T[\mathbf{W} \bm{h}_i \oplus \mathbf{W} \bm{h}_u]\right)\right)
    },
\end{equation}
% \end{small}\noindent
where $\mathbf{W}\in\mathbb{R}^{d\times d}$ denotes the weight matrix, $\mathbf{a}\in\mathbb{R}^{2d}$ is a learnable parameter, $\oplus$ is the concatenation operation and $\eta$ is the LeakyReLU nonlinearity.
Motivated by~\cite{li2019semi}, we restrict $\mathbf{W}$ to be a diagonal matrix to reduce computations, thus increasing the scalability of the model.
To stabilize the learning process of self-attention~\cite{velivckovic2018graph}, we perform $K$ ($K=2$) heads of independent attention of Eq.~(\ref{gat}) in parallel, and concatenate these features to obtain the structure embedding of entity $e_i$:
\begin{gather}
    \bm{h}^g_i = \bigoplus^{K}_{k=1} \textstyle\sigma\left(\sum_{j\in{\mathcal{N}_i}}\alpha^{k}_{ij}\bm{h}_j\right),
\end{gather}
where $\alpha^{k}_{ij}$ is the normalized attention coefficient computed by the $k$-th attention.
% and $\mathbf{W}^{k}$ is the corresponding $k$-th weight matrix.
In practice, we apply a two-layer GAT model to aggregate the neighborhood information within multiple hops, and use the output of the last GAT layer as the neighborhood structure embedding.
% Also, existing graph representation learning algorithms can be easily incorporated into M2GRL.

\subsubsection{Relation, Attribute, and Name Embeddings}

Because the vanilla GAT operates on unlabeled graphs, it is unable to properly model relational information in multi-relational KGs.
% To alleviate this issue, we follow \cite{yang2019aligning} and regard the relations of entity $e_i$ as bag-of-words features and adopt a simple feed-forward layer to map relation features into low-dimensional spaces.
To alleviate this issue, we follow \cite{yang2019aligning} and regard the relations of entity $e_i$ as bag-of-words feature and feed it into a simple feed-forward layer to obtain the relation embedding $\bm{h}^r_i$.
For the simplicity and consistency of MCLEA, we adopt the same approach for the attribute embedding $\bm{h}^a_i$ and the name embedding $\bm{h}^n_i$ for entity $e_i$.
Therefore, these embeddings are calculated as:
% \begin{equation}
%     \{\bm{h}^r_i,\bm{h}^a_i,\bm{h}^n_i\} = \mathbf{W}_{\{r,a,n\}}\{\bm{r}_i,\bm{a}_i,\bm{n}_i\}+\mathbf{b}_{\{r,a,n\}},
% \end{equation}
\begin{equation}
    \bm{h}^l_i = \mathbf{W}_l\bm{u}^l_i + \mathbf{b}_l,\ l\in\{r,a,n\},
\end{equation}
where $\mathbf{W}_l$ and $\mathbf{b}_l$ are learnable parameters, $\bm{u}^r_i$ is the bag-of-words relation feature, $\bm{u}^a_i$ is the bag-of-words attribute feature, and $\bm{u}^n_i$ is the name feature obtained by averaging the pre-trained GloVe~\cite{pennington2014glove} vectors of name strings.
To avoid the out-of-vocabulary challenges brought by the extensive proper nouns (e.g., person names) and the limited vocabulary of word vectors, we further incorporate the character bigrams~\cite{mao2021alignment} of entity names as auxiliary features for name embedding.

\subsubsection{Visual Embedding}

We adopt the pre-trained visual model (PVM), e.g., ResNet-152~\cite{he2016deep}, to learn visual embedding, shown as ``\textbf{Visual Encoder}'' in Figure~\ref{fig:model}.
% Concretely, we feed the image $v_i$ of entity $e_i$ into the PVM backbone and take the last layer's output before logits as the image feature.
% We feed the image  $v_i$ of entity $e_i$ into ResNet-152 and use the 2048-dimensional output of the global average pooling (pool5) after the final convolutional layer.
% The feature is sent through a feed-forward layer to get the visual embedding $\bm{h}^v_i\in\mathbb{R}^{d_v}$ for entity $e_i$:
We feed the image $v_i$ of entity $e_i$ into the pre-trained visual model and use the final layer output before logits as the image feature.
The feature is sent through a feed-forward layer to get the visual embedding:
\begin{equation}
    \bm{h}^v_i = \mathbf{W}_v\cdot \text{PVM}({v_i})+\mathbf{b}_v.
\end{equation}

\subsubsection{Joint Embedding}

% We integrate the multi-modal features into a single compact representation $\hat{\bm{h}}_i$ for entity $e_i$.
% The elaborately designed fusion module may introduce high computational costs and may lead to overfitting due to scarce seed alignments.
% between cross-graph entities.
% A simple weighted concatenation is implemented:
Next, we implement a simple weighted concatenation by integrating the multi-modal features into a single compact representation $\hat{\bm{h}}_i$ for entity $e_i$:
\begin{small}
\begin{equation}
    \hat{\bm{h}}_i = \bigoplus_{m\in\mathcal{M}}
    \left[
    \frac{
    \exp(w_m)
    }{
    \sum_{j\in \mathcal{M}}\exp(w_j)
    }
    \bm{h}^{m}_{i}
    \right],
    \label{emb:joint}
\end{equation}
\end{small}\noindent
where $\mathcal{M}=\{g,r,a,n,v\}$ and $w_m$ is a trainable attention weight for the modality of $m$. L2-normalization is performed on the input embeddings before the weighted concatenation.

% Note that we regard the fusion embedding as a special modality, which integrates the information of all modalities and also serves as a feature of similarity measures in the inference stage. 
% Note that we regard the joint embeddings as a special modality, called joint modality, which is learned along with other unimodalities.

The current joint embeddings are coarse-grained and there are no interactions between modalities. 
Therefore, two training strategies are designed for learning the dynamics within (intra-) and between (inter-) modalities.

\subsection{Contrastive Representation Learning}
\label{sec:cl}

% As the core of our proposed method, the two proposed losses operated on the encoded unimodal and joint representations are designed to perform intra-/inter-modal learning in the training stage. With the designed losses, MCLEA can sufficiently learn dynamics between and within modalities, preserve semantic proximity and minimize the modality gap.
As the core of MCLEA, we propose two novel losses on the uni-modal and joint representations to sufficiently capture the dynamics within and between modalities while preserving semantic proximity and minimizing the modality gap.

\subsubsection{Intra-modal Contrastive Loss (ICL)} 

% The principle behind contrastive learning (CL)~\cite{chen2020simple,khosla2020supervised} is to define positive and negative samples relative to an anchor point, where the goal is to pull together an anchor and a positive sample in the embedding space, and push apart the anchor from many negative samples.
% Inspired by the successful application of CL, we devise a contrastive objective to distinguish the embeddings of the same entity under the two views from the embeddings of other entities, so as to leverage the supervision signals in the unlabeled data.
% The principle behind contrastive learning (CL)~\cite{chen2020simple,khosla2020supervised} is to pull together an anchor and a positive sample in the embedding space, and push apart the anchor from many negative samples.
Inspired by recent work on contrastive learning (CL)~\cite{chen2020simple,khosla2020supervised}, we devise an intra-modal contrastive loss (ICL) that enforces the input embedding to respect the similarity of entities in the original embedding space.
Meanwhile, ICL allows MCLEA to distinguish the embeddings of the same entities in different KGs from those of other entities for each modality.

Given that $\mathcal{S}$ can be naturally regarded as positive samples, whereas any non-aligned pairs can be regarded as negative samples due to the convention of 1-to-1 alignment constraint~\cite{sun2018bootstrapping}. 
% all combinations $(x_i,y_j)$ and $(e^i_1,e^j_2)$ with $i\neq{j}$ can therefore serve as unaligned pairs.
Formally, for the $i$-th entity $e^i_1\in E_1$ of minibatch $\mathcal{B}$, the positive set is defined as ${P_i}=\{e^i_2|e^i_2\in E_2\}$, where $(e^i_1,e^i_2)$ is an aligned pair.
The negative set includes two types, inner-graph unaligned pairs from the source KG $G_1$ and cross-graph unaligned pairs from the target KG $G_2$, defined as $N^i_1=\{e^j_1|\forall{e}^j_1\in{E_1}, j\neq{i}\}$ and $N^i_2=\{e^j_2|\forall{e}^j_2\in{E_2},j\neq{i}\}$, respectively.
Both $N^i_1$ and $N^i_2$ come from minibatch $\mathcal{B}$.
% The purpose of including two types is not only to force the features from the two KGs map to proximate points in the joint embedding, but also to retain that similar features from the same KG stay close-by in the joint embedding.
These two types of negative samples are designed to constrain the joint embedding space, in which  the semantically similar entities from the same KG stay close-by and the aligned entities from two KGs map to proximate points. 
% Overall, we define the intra-modal contrastive loss of the $k$-th modality for each positive pair $(x_i,y_i)$ on a minibatch $\mathcal{M}$ as:
Overall, we define the alignment probability distribution $q_m(e^i_1,e^i_2)$ of the modality of $m$ for each positive pair $(e^i_1,e^i_2)$ as:
% \begin{gather}
%     \mathcal{L}_k = - \mathbb{E}_{i\in\mathcal{M}} \left[\log\frac{\exp(f_k(x_i)^T f_k(y_i)/\tau)}{\exp(f_k(x_i)^T f_k(y_i)/\tau) + {J}^{\text{inner}}_k + {J}^{\text{cross}}_k}\right] \\
%     {J}^{\text{cross}}_k = \sum_{y_j\in{N_i}}\exp(f_k(x_i)^T f_k(y_j)/\tau)
% \end{gather}
% \begin{gather}
% % \begin{small}
% % \begin{aligned}
%     % \ell_k(x_i,y_i) 
%     q_m(e^i_1,e^i_2)
%     = 
%     \frac{\delta_m{(e^i_1,e^i_2)}}{
%     \delta_m{(e^i_1,e^i_2)} +
%      {J}^{1}_m + 
%     {J}^{2}_m
%     }, \label{eq:icl}\\
%     {J}^{1}_m = \sum\limits_{e^j_1\in{N}^i_1}\delta_m(e^i_1,e^j_1), 
%     {J}^{2}_m = \sum\limits_{e^j_2\in{N}^i_2}\delta_m(e^i_1,e^j_2)
% % \end{aligned}
% % \end{small}
% \end{gather}
% \begin{gather}
%     q_k(y_i|x_i)
%     = 
%     \frac{\delta_k{(x_i,y_i})}{
%     \delta_k{(x_i,y_i)} +
%      {J}^{1}_k + 
%     {J}^{2}_k
%     }, \label{eq:icl}\\
%     {J}^{1}_k = \textstyle\sum_{x_j\in{N}^1_i}\delta_k(x_i,x_j), 
%     {J}^{2}_k = \textstyle\sum_{y_j\in{N}^2_i}\delta_k(x_i,y_j),
% \end{gather}
\begin{equation}
\begin{small}
\begin{aligned}
    & q_m(e^i_1,e^i_2)= \\
    & \frac{\delta_m{(e^i_1,e^i_2})}{
    \delta_m{(e^i_1,e^i_2)} +
    \sum\limits_{e^j_1\in{N}^i_1}\delta_m(e^i_1,e^j_1) + \sum\limits_{e^j_2\in{N}^i_2}\delta_m(e^i_1,e^j_2)
    % \textstyle\sum_{y_j\in{N}^2_i}\delta_k(x_i,y_j
    },
    \label{eq:icl}
\end{aligned}
\end{small}
\end{equation}
where $\delta_m(u,v)=\exp(f_m(u)^Tf_m(v)/\tau_1)$, $f_m(\cdot)$ is the encoder of the modality $m$, and $\tau_1$ is a temperature parameter.
% Similar to SCL, L2-normalization on the representation is performed such that the similarity of each pair is between 0 and 1.
Especially, L2-normalization is performed on the input feature embedding before computing the inner product.
% \todo{The negative samples come from two sources, inner-graph or cross-graph entity pairs, corresponding to the second and the third term in the denominator of Eq.~(\ref{eq:icl}), respectively.}
Notably, the distribution of Eq.~(\ref{eq:icl}) is directional and asymmetric for each input; the distribution for another direction is thus defined similarly for $q_m(e^i_2,e^i_1)$. 
% The overall objective to be maximized is then defined as the average over all positive pairs, formally given by
% where $\mathbb{E}$ is an expectation operator over all positive pairs in a mini-batch.
The ICL can be formulated as:
% \begin{small}
\begin{equation}
\begin{small}
\begin{aligned}
    \mathcal{L}^{\text{ICL}}_m = - \mathbb{E}_{i\in\mathcal{B}}
    \log
    \left[
    \frac{1}{2}
    (q_m(e^i_1,e^i_2) + q_m(e^i_2,e^i_1))
    \right].
\end{aligned}
\end{small}
\end{equation}
% \end{small}\noindent

We apply ICL on each modality separately and also on the combined multi-modal representation as specified in Eq.~(\ref{emb:joint}).
ICL is performed in contrastive supervised learning to learn intra-modal dynamics for more discriminative boundaries for each modality in the embedding space. 
% ICL encourages the representations of the align pairs from same modality to have the highest similarity, and forces the representations of the unalign pairs from same modality to have the lowest similarity.

\subsubsection{Inter-modal Alignment Loss (IAL)}

Since the embeddings of different modalities are separately trained with ICL, their representations are not aligned and it is difficult to model the complex interaction between modalities solely with the fusion module.
% Therefore, we further propose to utilize the fusion embedding with stronger representation to improve the accuracy of textual input view, based on cross-view alignment that minimizes the KL divergence over the probability distributions of the two modalities.
To alleviate this, we further propose inter-modal alignment loss (IAL), which targets at reducing the gap between the output distribution over different modalities, so that the MCLEA can model inter-modal interactions and obtain more meaningful representations.

% We notice that the joint embedding has a comprehensive representation due to the fusion of multiple modal features, so we transfer the knowledge from joint embedding to uni-modal embedding so that uni-modality can better utilize the complementary information from others.
We consider the joint embedding as the comprehensive representation due to its fusion of multi-modal features; therefore, we attempt to transfer the knowledge from the joint embedding back to uni-modal embedding so that the uni-modal embedding could better utilize the complementary information from others.
Concretely, we minimize the bidirectional KL divergence over the output distribution between joint embedding and uni-modal embedding:
% \begin{gather}
% \begin{aligned}
%     \mathcal{L}^{\text{IAL}}_k
%     &= \mathbb{E}_{i\in\mathcal{B}}
%     \text{KL}(q_o(x_i,y_i)||q_k(x_i,y_i)) \\
%     &=\mathbb{E}_{i\in\mathcal{B}} 
%     \left[
%     q_o(x_i,y_i)\log\frac{q_o(x_i,y_i)}{q_k(x_i,y_i)}
%     \right],
% \end{aligned}
% \end{gather}
% \begin{small}
\begin{equation}
\begin{small}
\begin{aligned}
    \mathcal{L}^{\text{IAL}}_m
    = 
    \mathbb{E}_{i\in\mathcal{B}}\ 
    \frac{1}{2}
    [
    &\text{KL}(q'_o(e^i_1,e^i_2)\ ||\ q'_m(e^i_1,e^i_2)) \\
    \quad+ &\text{KL}(q'_o(e^i_2,e^i_1)\ ||\ q'_m(e^i_2,e^i_1))
    ],
    % &=\mathbb{E}_{i\in\mathcal{B}} 
    % \left[
    % q_o(x_i,y_i)\log\frac{q_o(x_i,y_i)}{q_k(x_i,y_i)}
    % \right]
    \label{eq:ial}
\end{aligned}
\end{small}
\end{equation}
% \end{small}\noindent
where $q'_o(e^i_1,e^i_2)$, $q'_o(e^i_2,e^i_1)$ and $q'_m(e^i_1,e^i_2)$, $q'_m(e^i_2,e^i_1)$ represent the output predictions with two directions of joint embedding and the uni-modal embedding of modality $m$, respectively.
% They are calculated in the same way as Eq.~(\ref{eq:icl}) but with a temperature parameter $\tau_2$.
Their calculation are similar to Eq.~(\ref{eq:icl}) but with a temperature parameter $\tau_2$.
We only back-propagate through $q'_m(e^i_1,e^i_2)$, $q'_m(e^i_2,e^i_1)$ in Eq.~(\ref{eq:ial}) as knowledge distillation~\cite{hinton2015distilling}.

IAL aims at learning interactions between different modalities within each entity, which concentrates on aggregating the distribution of different modalities and thus reduces the modality gap.
Some approaches~\cite{zhang2019multi,chen2020mmea} attempt to learn a common space by imposing alignment constraints on the features between different modalities, but they introduce noises due to semantic heterogeneity. 
Different from these approaches, we distill the useful knowledge from the output prediction of multi-modal representation to the uni-modal representation, while maintaining relatively modality-specific features of each modality.
% However, we align the output distribution, similar to knowledge distillation, which has merged the learning of different features, Allowing each modality to absorb complementary information from the other modalities.

\subsection{Optimization Objective}
\label{sec:objective}
% MuCRL is concerned about learning a global optimization with respect to ICL and IAL for each modality.
% MuCRL can also be viewed as multi-task learning.
The overall loss of the MCLEA is as follows,
% \begin{small}
\begin{equation}
\begin{small}
\begin{aligned}
    \mathcal{L} = \mathcal{L}^{\text{ICL}}_o + 
    \textstyle\sum_{m\in\mathcal{M}} \alpha_m\mathcal{L}^{\text{ICL}}_m + 
    \textstyle\sum_{m\in\mathcal{M}} \beta_m\mathcal{L}^{\text{IAL}}_m,
    \label{eq:loss}
\end{aligned}
\end{small}
\end{equation}
% \end{small}\noindent
where $\mathcal{M}=\{g,r,a,n,v\}$, $\mathcal{L}^{\text{ICL}}_o$ denotes the ICL operated on joint embedding, $\alpha_m$ and $\beta_m$ are hyper-parameters that balance the importance of losses.
However, manually tuning these hyper-parameters is expensive and intractable.
Instead, we treat MCLEA as a multi-task learning paradigm and then use homoscedastic uncertainty~\cite{kendall2018multi} to weigh each loss automatically during model training.
We adjust the relative weight of each task in the loss function by deriving a multi-task loss function based on maximizing the Gaussian likelihood with task-dependant uncertainty.
% Instead, inspired by the recent study which uses uncertainty to weigh losses in multi-task learning [11], we leverage a Bayesian task weight learner that can automatically achieve the balance among multi-tasks. Due to space limit, here we only show the derived result and put the detailed derivation process in the Supplement A.5.
Due to space limits, we only show the derived result and leave the detailed derivation process in the Appendix. The loss in Eq.~(\ref{eq:loss}) can be rewritten as:
% \begin{equation}
% \begin{aligned}
%     \mathcal{L} 
%     = \mathcal{L}^{\text{ICL}}_o &+ 
%     \textstyle\sum_{m\in\mathcal{M}} \left(\frac{1}{\alpha^2_m}\mathcal{L}^{\text{ICL}}_m +\log{\alpha_m}\right) \\
%     \qquad&+ 
%     \textstyle\sum_{m\in\mathcal{M}} \left(\frac{1}{\beta^2_m}\mathcal{L}^{\text{IAL}}_m +\log{\beta_m}\right),
%     \label{eq:loss2}
% \end{aligned}
% \end{equation}
\begin{equation}
\begin{small}
\begin{aligned}
    \mathcal{L} 
    = \mathcal{L}^{\text{ICL}}_o + 
    \textstyle\sum\limits_{m\in\mathcal{M}} 
    \left(\frac{1}{\alpha^2_m}\mathcal{L}^{\text{ICL}}_m + 
    \frac{1}{\beta^2_m}\mathcal{L}^{\text{IAL}}_m + 
    \log{\alpha_m} + \log{\beta_m}\right),
    \label{eq:loss2}
\end{aligned}
\end{small}
\end{equation}
where $\alpha_m$ and $\beta_m$ are automatically learned during training.
% where $\alpha_m$ and $\beta_m$ are trainable parameters.

To overcome the lack of training data, we incorporate a bi-directional iterative strategy used in~\cite{liu2021visual} to iteratively add new aligned seeds during training.
In the inference, we use the cosine similarity metric between joint embeddings of entities to determine the counterparts of entities.

% A few methods have investigated the unsupervised alignment that mine semantic similarities from the entity names~\cite{mao2020mraea,ge2021make} or images~\cite{liu2021visual} and use them to produce preliminary pseudo-labeled data.
% Our method can also be naturally extended to an unsupervised setting where entity pairs are discovered based on feature similarities of names or images as pseudo-alignment seeds for training our model.

MCLEA can be extended to the unsupervised setting, in which the pseudo-alignment seeds are discovered based on feature similarities of entity names~\cite{mao2020mraea,ge2021make} or entity images~\cite{liu2021visual}, accordingly resulting in different unsupervised versions of MCLEA.

\section{Experiments}

\subsection{Experimental Setup}
\noindent\textbf{Datasets}.
Five EA datasets are adopted for evaluation, including three bilingual datasets ZH/JA/FR-EN versions of DBP15K~\cite{liu2021visual} and two cross-KG datasets FB15K-DB15K/YAGO15K~\cite{liu2019mmkg}.
The detailed dataset statistics are listed in Table~\ref{tab:dataset} in the Appendix.
Note that not all entities have corresponding images and for those without images, MCLEA would assign random vectors for the visual modality, as the setting of EVA~\cite{liu2021visual}.
As for DBP15K, 30\% reference entity alignments are given as $\mathcal{S}$ while as for cross-KG datasets, 20\%, 50\%, and 80\% reference entity alignments are given~\cite{liu2019mmkg}.

\noindent\textbf{Baselines}.
% we compare MCLEA against state-of-the-art baselines in entity alignment, including three types of models.
We compare the proposed MCLEA against 19 state-of-the-art EA methods, which can be classified into four categories:
1) \textit{structure-based methods} that solely rely on structural information for aligning entities, including BootEA~\cite{sun2018bootstrapping}, MUGNN~\cite{cao2019multi}, KECG~\cite{li2019semi}, NAEA~\cite{zhu2019neighborhood}, and AliNet~\cite{sun2020knowledge};
2) \textit{auxiliary-enhanced methods} that utilize auxiliary information to improve the performance, including MultiKE~\cite{zhang2019multi}, HMAN~\cite{yang2019aligning},
RDGCN~\cite{wu2019relation}, AttrGNN~\cite{liu2020exploring}, BERT-INT~\cite{tang2021bert} and ERMC~\cite{yang2021entity};
3) \textit{multi-modal methods} that combine the multi-modal features to generate entity representations, including PoE~\cite{liu2019mmkg}, MMEA~\cite{chen2020mmea}, HMEA~\cite{guo2021multi}, and EVA~\cite{liu2021visual};
4) \textit{unsupervised methods}, including RREA~\cite{mao2020relational}, MRAEA~\cite{mao2020mraea}, EASY~\cite{ge2021make}, and SEU~\cite{mao2021alignment}.

\noindent\textbf{Implementation Details}.
% If entities’ names are in different languages in two KGs, following previous work~\cite{xu2019cross}, we use machine translation tool to translate names in non-English language to English language.
% For bilingual DBP15K, we use machine translation tool to translate entity names from both languages into the same language.
The hidden size of each layer of GAT is 300, while the embedding size of the other modalities is 100. We use the AdamW optimizer with a learning rate of $5 \times {10}^{-4}$ to update the parameters. The number of training epochs is 1000 with early-stopping and the batch size is 512. The hyper-parameters $\tau_1, \tau_2$ are set to 0.1 and 4.0, respectively. 
% To make fair comparisons with previous work, we use the same entity name translations and word vectors provided by~\cite{xu2019cross} for bilingual datasets.
To keep in line with previous works, we use the same entity name translations and word vectors provided by~\citet{xu2019cross} for bilingual datasets.
As for cross-KG datasets, we do not consider surface forms for fair comparison.
For visual embedding, we adopt the preprocessed image features provided by~\citet{liu2021visual} and~\citet{chen2020mmea} for bilingual datasets and cross-KG datasets, where the former uses ResNet-152 as the pre-trained backbone, while the latter uses VGG-16.
Previous work has revealed that surface forms are quite helpful for entity alignment~\cite{liu2020exploring}. For fair comparison, we divide the supervised methods on bilingual datasets into two groups based on whether surface forms are used, and we implement an MCLEA variant (MCLEA$\dagger$) where the name embedding is removed.
For the unsupervised setting, we implement two variants, MCLEA-V and MCLEA-N, which generate pseudo-alignment seeds based on the similarities of images and names, respectively.

\noindent\textbf{Evaluation}.
% Following the convention~\cite{liu2021visual},
We rank matching candidates of each to-be-aligned entity and use the metrics of Hits@1 (H@1), Hits@10 (H@10), and mean reciprocal rank (MRR).
In the following tables, the best results are \textbf{in bold} with the second best results \underline{underlined}, and ``Improv. best \%'' denotes the relative improvement of MCLEA over the best baseline.

\begin{table*}[ht]
    \centering
    \footnotesize
    \renewcommand\arraystretch{1.0}
    \begin{tabular}{@{}l|l|ccc|ccc|ccc@{}}
        \toprule
        & \multirow{2}*{Models} & \multicolumn{3}{c|}{DBP15K$_{ZH-EN}$} & \multicolumn{3}{c|}{DBP15K$_{JA-EN}$} & \multicolumn{3}{c}{DBP15K$_{FR-EN}$} \\
        & & {\scriptsize H@1} & {\scriptsize H@10} & {\scriptsize MRR} & {\scriptsize H@1} & {\scriptsize H@10} & {\scriptsize MRR} & {\scriptsize H@1} & {\scriptsize H@10} & {\scriptsize MRR} \\
        \midrule
        \parbox[t]{2mm}{\multirow{7}{*}{\rotatebox[origin=c]{90}{w/o SF}}} & BootEA {\scriptsize \cite{sun2018bootstrapping}} & 
        .629 & .847 & .703 & .622 & .854 & .701 & .653 & .874 & .731 \\
        & KECG {\scriptsize \cite{li2019semi}} & 
        .478 & .835 & .598 & .490 & .844 & .610 & .486 & .851 & .610 \\
        & MUGNN {\scriptsize \cite{cao2019multi}} & 
        .494 & .844 & .611 &  .501 & .857 & .621 & .495 & .870 & .621 \\
        & NAEA {\scriptsize \cite{zhu2019neighborhood}} & 
        .650 & .867 & .720 & .641 & .873 & .718 & .673 & .894 & .752 \\
        & AliNet {\scriptsize \cite{sun2020knowledge}} & 
        .539 & .826 & .628 & .549 & .831 & .645 & .552 & .852 & .657 \\
        % \cmidrule(lr){2-11}
        & EVA {\scriptsize \cite{liu2021visual}} & 
        \underline{.761} & \underline{.907} & \underline{.814} & \underline{.762} & \underline{.913} & \underline{.817} & \underline{.793} & \underline{.942} & \underline{.847} \\
        % \cmidrule(lr){2-11}
        & MCLEA$\dagger$ {\scriptsize (Ours)} & 
        \textbf{.816} & \textbf{.948} & \textbf{.865} & \textbf{.812} & \textbf{.952} & \textbf{.865} & \textbf{.834} & \textbf{.975} & \textbf{.885} \\
        \cmidrule(lr){2-11}
        & Improv. best \% &
        7.2 & 4.5 & 6.3 & 6.6 & 4.3 & 5.9 & 5.2 & 3.5 & 4.5 \\
        \midrule
        \parbox[t]{2mm}{\multirow{7}{*}{\rotatebox[origin=c]{90}{w/ SF}}} & MultiKE {\scriptsize \cite{zhang2019multi}} & 
        .437 & .516 & .466 & .570 & .643 & .596 & .714 & .761 & .733 \\
        & HMAN {\scriptsize \cite{yang2019aligning}} & 
        .562 & .851 & -- & .567 & .969 & -- & .540 & .871 & -- \\
        & RDGCN {\scriptsize \cite{wu2019relation}} & 
        .708 & .846 & -- & .767 & .895 & -- & .886 & .957 & -- \\
        & AttrGNN {\scriptsize \cite{liu2020exploring}} & 
        .777 & .920 & .829 & .763 & .909 & .816 & .942 & .987 & .959 \\
        & BERT-INT {\scriptsize \cite{tang2021bert}} & 
        \underline{.968} & \underline{.990} & \underline{.977} & \underline{.964} & \underline{.991} & \underline{.975} & \underline{.992} & \underline{.998} & \underline{.995} \\
        & ERMC {\scriptsize \cite{yang2021entity}} & 
        .903 & .946 & .899 & .942 & .944 & .925 & .962 & .982 & .973 \\
        & MCLEA {\scriptsize (Ours)} & 
        \textbf{.972} & \textbf{.996} & \textbf{.981} & \textbf{.986} & \textbf{.999} & \textbf{.991} & \textbf{.997} & \textbf{1.00} & \textbf{.998} \\
        \cmidrule(lr){2-11}
        & Improv. best \% &
        0.4 & 0.6 & 0.4 & 2.3 & 0.8 & 1.6 & 0.5 & 0.2 & 0.3 \\
        \bottomrule
    \end{tabular}
    % \caption{Experimental results on three bilingual datasets where SF denotes the surface forms.}
    \caption{Comparative results of MCLEA without (w/o) and with (w/) surface forms (SF) against strong supervised methods on three bilingual datasets, and $\dagger$ denotes an MCLEA variant without name embedding.}
    \label{tab:overall-perf-1}
\end{table*}

\subsection{Overall Results}

% Table~\ref{tab:overall-perf-1} and Table~\ref{tab:overall-perf-un} report the performance of MCLEA against other baselines on bilingual datasets in both supervised and unsupervised settings, respectively. Table~\ref{tab:overall-perf-2} shows the performance of MCLEA compared to other multi-modal methods with different alignment seeds.

Table~\ref{tab:overall-perf-1}, Table~\ref{tab:overall-perf-2}, and Table~\ref{tab:overall-perf-un} report the performance of MCLEA against different baselines on different datasets with different settings.
Overall, MCLEA and its variants mostly perform the best across all the datasets on all the metrics.

% As shown in Table~\ref{tab:overall-perf-1}, we find that, in the supervised setting, MCLEA greatly outperforms all the structure-based, auxiliary-enhanced, and multi-modal methods except BERT-INT by a large margin from 5\% to 20\% in H@1.
% BERT-INT achieves near-perfect scores as it directly fine-tunes a multi-lingual BERT as the backbone; MCLEA shows slightly better performance than BERT-INT nevertheless with far fewer parameters (13M $vs.$ 110M).
Table~\ref{tab:overall-perf-1} reports the performance of MCLEA against the supervised baselines on bilingual datasets in the settings of w/ and w/o surface forms.
Compared with the first group without using surface forms, MCLEA$\dagger$ brings about 5.2\% to 7.2\% relative improvement in H@1 over the best baseline EVA.
The superiority of MCLEA confirms that the proposed contrastive representation learning substantially promotes the performance.
Specifically, compared with the second group with surface forms involvement, there are two notable observations.
On one hand, MCLEA shows a clear improvement when combined with name embedding, suggesting that entity names provide useful clues for entity alignment, which has been revealed in previous work~\cite{zhang2019multi,liu2020exploring, ge2021make}.
On the other hand, MCLEA still shows slightly better performance than the best baseline BERT-INT with 0.4\% to 2.3\% relative improvement in H@1 nevertheless with far fewer parameters (13M $vs.$ 110M).
This also reveals that MCLEA can effectively model robust entity representations instead of attaching over-parameterized encoders.
Noteworthily, BERT-INT relies heavily on entity descriptions to fine-tune BERT, but entity descriptions may not be available for every entity, and collecting them is labor-intensive, limiting the scope of its application.

Table~\ref{tab:overall-perf-2} shows the comparison of multi-modal methods on two cross-KG datasets, which provides direct evidence of the effectiveness of MCLEA.
When 20\% training seeds are given, MCLEA outperforms the best baseline MMEA with 67.9\% higher in H@1, 30.3\% higher in H@10, and 49.6\% higher in MRR.
The performance gains are still significant when 50\% and 80\% alignment seeds are given.
It is worth noting that the performance gains reach the highest in the 20\% setting and MCLEA (20\%) obtains comparable results to EVA (80\%), indicating that MCLEA could better utilize the minimum number of alignment seeds to obtain effective representations.
We also find that MMEA greatly outperforms EVA, we speculate that the cross-KG datasets are quite heterogeneous (w.r.t. the number of relations) compared to bilingual datasets, as shown in Table \ref{tab:dataset} in the Appendix, and the structure encoder of EVA struggles to model heterogeneous information and EVA cannot utilize the numerical knowledge in cross-KG datasets, which is well exploited in MMEA.

\begin{table}[ht]
    \centering
    \scriptsize
    \renewcommand\arraystretch{1.0}
    \begin{tabular}{@{}l|l|ccc|ccc@{}}
        \toprule
        & \multirow{2}*{Models} & \multicolumn{3}{c|}{FB15K-DB15K} & \multicolumn{3}{c}{FB15K-YAGO15K} \\
        & & {\tiny H@1} & {\tiny H@10} & {\tiny MRR} & {\tiny H@1} & {\tiny H@10} & {\tiny MRR} \\
        \midrule
        \parbox[t]{0.5mm}{\multirow{5}{*}{\rotatebox[origin=c]{90}{20\%}}} & PoE  & 
        .126 & .251 & .170 & .113 & .229 & .154 \\
        & HMEA & .127 & .369 & -- & .105 & .313 & -- \\
        & MMEA & \underline{.265} & \underline{.541} & \underline{.357} & \underline{.234} & \underline{.480} & \underline{.317} \\
        & EVA$\ast$ & .134 & .338 & .201 & .098 & .276 & .158 \\
        & MCLEA {\tiny(Ours)} & \textbf{.445} & \textbf{.705} & \textbf{.534} & \textbf{.388} & \textbf{.641} & \textbf{.474} \\
        % & \textbf{MCLEA} & \textbf{.992} & \textbf{.998} & \textbf{.995} & \textbf{.992} & \textbf{.996} & \textbf{.994} \\
        \cmidrule(lr){2-8}
        & Improv. best \% & 
        67.9 & 30.3 & 49.6 & 65.8 & 33.5 & 49.5 \\
        \midrule
        \parbox[t]{0.5mm}{\multirow{5}{*}{\rotatebox[origin=c]{90}{50\%}}} & PoE & 
        \underline{.464} & .658 & \underline{.533} & .347 & .536 & .414 \\
        & HMEA & .262 & .581 & -- & .265 & .581 & -- \\
        & MMEA & .417 & \underline{.703} & .512 & \underline{.403} & \underline{.645} & \underline{.486} \\
        & EVA$\ast$ & .223 & .471 & .307 & .240 & .477 & .321 \\
        & MCLEA {\tiny(Ours)} & \textbf{.573} & \textbf{.800} & \textbf{.652} & \textbf{.543} & \textbf{.759} & \textbf{.616} \\
        % & \textbf{MCLEA} & \textbf{.993} & \textbf{.998} & \textbf{.995} & \textbf{.994} & \textbf{.998} & \textbf{.995} \\
        \cmidrule(lr){2-8}
        & Improv. best \% & 
        23.5 & 13.8 & 22.3 & 34.7 & 17.7 & 26.7 \\
        \midrule
        \parbox[t]{0.5mm}{\multirow{5}{*}{\rotatebox[origin=c]{90}{80\%}}} & PoE & 
        \underline{.666} & .820 & \underline{.721} & .573 & .746 & .635 \\
        & HMEA & .417 & .786 & -- & .433 & .801 & -- \\
        & MMEA & .590 & \underline{.869} & .685 & \underline{.598} & \textbf{.839} & \underline{.682} \\
        & EVA$\ast$ & .370 & .585 & .444 & .394 & .613 & .471 \\
        & MCLEA {\tiny(Ours)} & \textbf{.730} & \textbf{.883} & \textbf{.784} & \textbf{.653} & \underline{.835} & \textbf{.715} \\
        % & \textbf{MCLEA} & \textbf{.996} & \textbf{.999} & \textbf{.997} & \textbf{.995} & \textbf{.997} & \textbf{.996} \\
        \cmidrule(lr){2-8}
        & Improv. best \% & 
        9.6 & 1.6 & 8.7 & 9.2 & -0.4 & 4.8 \\
        \bottomrule
    \end{tabular}
    \caption{Experimental results on two cross-KG datasets where \textit{X}\% represents the percentage of reference entity alignments used for training. The symbol $\ast$ denotes the reproduced results.
    }
    \label{tab:overall-perf-2}
\end{table}

When compared to the unsupervised methods in Table \ref{tab:overall-perf-un}, both MCLEA variants perform slightly better than the best baseline with performance gain in H@1 varying from 0.7\% to 6.7\%.
Note that using image (-V) or name (-N) similarities to produce seeds leads to almost identical results, demonstrating the effectiveness of such simple rules to enable MCLEA in the unsupervised setting.

% Table~\ref{tab:overall-perf-2} shows the comparison of multi-modal methods on two cross-KG datasets, which provides direct evidence of the effectiveness of MCLEA.
% When 20\% alignment seeds are given for training, MCLEA outperforms the best baseline MMEA more than 270\% higher in H@1, 80\% higher in H@10, and 170\% higher in MRR.
% Since we notice that in two cross-KG datasets, equivalent entities typically share similar names, we implement an MCLEA variant, MCLEA$\dagger$, without name embedding to rule out this discriminative feature.
% For MCLEA$\dagger$, the performance gains are still obvious, more than 60\% in H@1, 30\% in H@10, and 40\% in MRR.
% Similar findings are also observed when 50\% and 80\% alignment seeds are given.
% It is worth noting that the performance gains reach the highest in the 20\% setting and MCLEA$\dagger$ (20\%) obtains comparable results to EVA (80\%), indicating that MCLEA could better utilize the minimum number of alignment seeds to obtain effective representations.

\begin{table*}[ht]
    \centering
    \footnotesize
    \renewcommand\arraystretch{1.0}
    \begin{tabular}{l|ccc|ccc|ccc@{}}
        \toprule
        \multirow{2}*{Models} & \multicolumn{3}{c|}{DBP15K$_{ZH-EN}$} & \multicolumn{3}{c|}{DBP15K$_{JA-EN}$} & \multicolumn{3}{c}{DBP15K$_{FR-EN}$} \\
        & {\scriptsize H@1} & {\scriptsize H@10} & {\scriptsize MRR} & {\scriptsize H@1} & {\scriptsize H@10} & {\scriptsize MRR} & {\scriptsize H@1} & {\scriptsize H@10} & {\scriptsize MRR} \\
        \midrule
        MRAEA {\scriptsize \cite{mao2020mraea}} & 
        .778 & .832 & -- & .889 & .927 & -- & .950 & .970 & -- \\
        RREA {\scriptsize \cite{mao2020relational}} &
        .822 & .964 & -- & .918 & .978 & -- & .963 & .992 & -- \\
        EASY {\scriptsize \cite{ge2021make}} & 
        .898 & \underline{.979} & \underline{.930} & .943 & .990 & .960 & .980 & .998 & .990 \\
        SEU {\scriptsize \cite{mao2021alignment}} & 
        \underline{.900} & .965 & .924 & \underline{.956} & \underline{.991} & \underline{.969} &  \underline{.988} & \underline{.999} & \underline{.992}\\
        % \cmidrule(lr){2-10}
        MCLEA-V {\scriptsize (Ours)} & 
        .959 & \textbf{.995} & \textbf{.974} & .977 & \textbf{.999} & .987 & .990 & \textbf{1.00} & .994 \\
        MCLEA-N {\scriptsize (Ours)} & \textbf{.960} & 
        .994 & \textbf{.974} & \textbf{.983} & \textbf{.999} & \textbf{.990} & \textbf{.995} & \textbf{1.00} & \textbf{.997} \\
        % \cmidrule(lr){1-10}
        \midrule
        Improv. best \% &
        6.7 & 1.6 & 4.7 & 2.8 & 0.8 & 2.2 & 0.7 & 0.1 & 0.5 \\
        \bottomrule
    \end{tabular}
    \caption{Unsupervised experimental results on three bilingual datasets, where -V and -N denote different methods to generate pseudo-alignment seeds.}
    \label{tab:overall-perf-un}
\end{table*}

\subsection{Model Analysis}

\noindent\textbf{Ablation study.}
The ablation experiments are performed on two bilingual datasets and the results are presented in Table~\ref{tab:ablation}.
% The ablation study is performed to provide insights into different components of MCLEA, and the results on two bilingual datasets are presented in Table~\ref{tab:ablation}.
We first examine the individual contribution of different modalities.
The removal of different modalities has varying degrees of performance drop, and entity names have shown the primary importance with the most significant drop, which is in line with the previous findings~\cite{mao2020mraea,ge2021make}.
The structural information also shows its stable effectiveness across different datasets and other modalities make a slight contribution to MCLEA. Especially, visual information can play a more pronounced role in the absence of surface forms~\cite{chen2020mmea, liu2021visual}.
% The structural information also shows its stable effectiveness across different datasets, while other modalities only make a minor contribution to MCLEA.
Furthermore, we inspect various training strategies in MCLEA.
It dramatically degrades the performance when removing the ICL from MCLEA, which indicates the importance of ICL in learning the intra-modal proximity.
% The removal of IAL also consistently (though slightly) decreases the performance, revealing the learning of interdependence between different modalities contributes to model performance.
The IAL learns the interdependence between different modalities and is also beneficial to our model.
Training MCLEA without the iterative strategy and replacing the uncertainty mechanism with uniform weights (i.e., w/o uncertainty) also cause decreases in performance.
Overall, the ablation experiments validate the involvement of these modalities and training strategies with empirical evidence.

\begin{table}[h]
    \centering
    \scriptsize
    \renewcommand\arraystretch{1.0}
    \begin{tabular}{@{}l|l|ccc|ccc@{}}
        \toprule
        & \multirow{2}*{Models} & \multicolumn{3}{c|}{DBP15K$_{ZH-EN}$} & \multicolumn{3}{c}{DBP15K$_{JA-EN}$} \\
        & & {\tiny H@1} & {\tiny H@10} & {\tiny MRR} & {\tiny H@1} & {\tiny H@10} & {\tiny MRR} \\
        \midrule
        & MCLEA & \textbf{.972} & \textbf{.996} & \textbf{.981} & \textbf{.986} & \textbf{.999} & \textbf{.991} \\
        \midrule
        \parbox[t]{1mm}{\multirow{5}{*}{\rotatebox[origin=c]{90}{Modalities}}} 
        & \textit{w/o} structure & .883 & .956 & .909 & .947 & .980 & .959 \\
        & \textit{w/o} relation & .967 & .995 & .978 & .985 & .999 & .991 \\
        & \textit{w/o} attribute & .961 & .994 & .974 & .983 & .999 & .991 \\
        & \textit{w/o} name & .816 & .948 & .865 &  .812 & .952 & .865 \\
        & \textit{w/o} visual & .968 & .994 & .978 & .985 & .999 & .991 \\
        \midrule
        \parbox[t]{1mm}{\multirow{4}{*}{\rotatebox[origin=c]{90}{Training}}} 
        & \textit{w/o} ICL & .782 & .892 & .818 & .813 & .909 & .844 \\
        & \textit{w/o} IAL & .966 & .995 & .977 & .980 & .998 & .987 \\
        & \textit{w/o} iter. strategy & .942 & .991 & .960 & .964 & .995 & .976 \\
        & \textit{w/o} uncertainty & .969 & .996 & .980 & .984 & .999 & .990 \\
        \bottomrule
    \end{tabular}
    \caption{Ablation study on two bilingual datasets.}
    \label{tab:ablation}
\end{table}

\begin{figure}[h]
    \centering
    \includegraphics[width=1.0\linewidth]{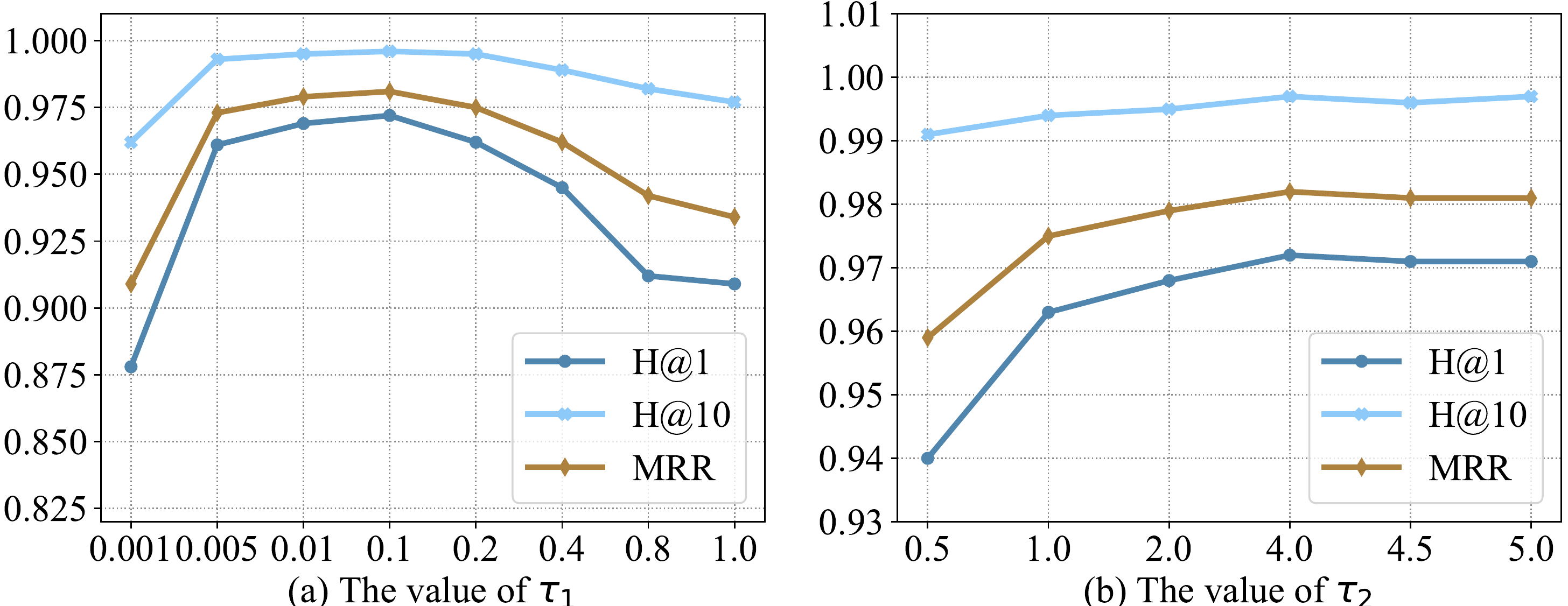}
    \caption{Performance comparison with different values of $\tau_1, \tau_2$.}
    \label{fig:params}
\end{figure}

\noindent\textbf{Impact of hyper-parameters $\tau_1,\tau_2$}.
We investigate the effects of hyper-parameters $\tau_1,\tau_2$ on $\text{DBP15K}_{ZH-EN}$.
As shown in Figure~\ref{fig:params}, different values of $\tau_1$ have drastic effects on MCLEA, especially in terms of H@1 and MRR, which is because $\tau_1$ controls the strength of penalties on hard negative samples and an appropriate $\tau_1$ is conducive to learning discriminative entity embeddings.
% On the other hand, \todo{$\tau_2$ has more robust effects}, and the performance reaches saturation point when $\tau_2=4.0$.
On the other hand, we observe lower variance in the performance w.r.t. $\tau_2$ and the performance saturates when $\tau_2=4.0$.
% $\tau_2$ regulates the softness of the alignment distribution produced by input embedding, thus establishing the associations between different modalities and transferring the generalization capability of the joint embedding to uni-modal embedding.
The KL divergence establishes the associations between different modalities, within which $\tau_2$ regulates the softness of the alignment distribution produced by input embedding and transfers the generalization capability of the joint embedding to uni-modal embedding.

\begin{figure}[h]
    \centering
    \includegraphics[width=1.0\linewidth]{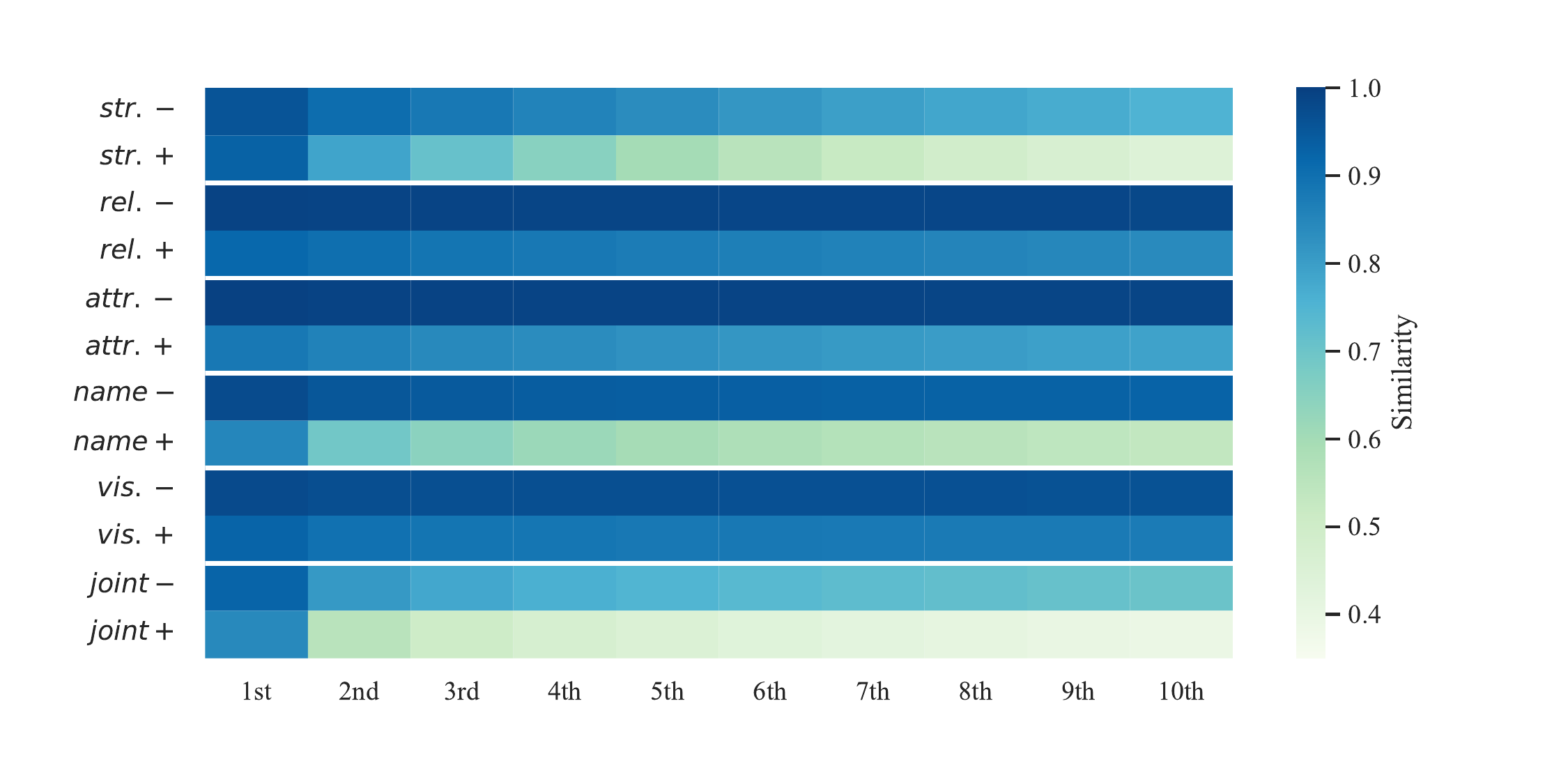}
    \caption{Similarity visualization of representations of test entities and their top-10 predicted counterparts. The vertical axis represents different modalities with ($+$) and without ($-$) ICL/IAL and the horizontal axis represents the index of ranked predictions.}
    \label{fig:sim-dist}
\end{figure}

\noindent\textbf{Similarity Distribution of Representations.}
% we experiment MCLEA with and without ICL/IAL on DBP15K$_{ZH-EN}$ and visualize the similarity distribution of the entity and its predicted counterparts for different modalities in Figure~\ref{fig:sim-dist}.
To investigate the effectiveness of entity representations, we experiment MCLEA with and without ICL/IAL on DBP15K$_{ZH-EN}$ and produce the visualization in Figure~\ref{fig:sim-dist} by averaging the similarity distribution of the test entities and their predicted counterparts for different modalities.
It can be observed that in every modality, especially in structure and name, it holds a high top-1 similarity and a large similarity variance.
More importantly, it meets our expectation that contrastive learning (ICL and IAL) enables more discriminative entity learning in the joint representations.

 \section{Conclusion}

% In this work, we learn representations using contrastive learning by leveraging abundant modalities naturally present in multi-modal knowledge graph.
% To this end, we propose a novel model MCLEA to address EA, which simultaneously considers intra-/inter-modal interactions innovated by contrastive learning. The model can ingest multiple modalities as input by simply adapting the encoder.
% MCLEA is composed of two types of objectives, i.e., intra-modal contrastive loss and inter-modal alignment loss.
% The ICL learns the representations of nodes within a single modality, and the IAL models the cross-modal relations of two different modalities.
% As MCLEA tends to have many optimization objectives, we adopt homoscedastic uncertainty to balance multiple objectives automatically.
% Equipped with these representations, we obtain state-of-the-art performance on multiple challenging benchmarks when compared to previous work.

This paper presented a novel method termed MCLEA to address the multi-modal entity alignment.
MCLEA utilizes multi-modal information to obtain the joint entity representations and it is composed of two losses, intra-modal contrastive loss, and inter-modal alignment loss, to explore the intra-modal relationships and cross-modal interactions, respectively.
We experimentally validated the state-of-the-art performance of MCLEA in several public datasets and its capability of learning more discriminative embedding space for entity alignment.
For future work, we plan to explore more side information such as entity descriptions to boost alignment performance.

% We also plan to extend the multi-modal contrastive learning framework to fit with image set setting~\cite{chen2020mmea}.

% \section*{Acknowledgments}

\bibliographystyle{acl_natbib}
\balance
\bibliography{anthology}

\newpage
\clearpage
\appendix
\section*{Appendix}
\label{sec:appendix}

\section{Derivation for Adaptively Weighted Multi-task Loss}

\begin{table*}[bp]
    \centering
    \footnotesize
    \renewcommand\arraystretch{1.0}
    \begin{tabular}{@{}l|c|rrrrrrrr@{}}
        \toprule
        Dataset & KG & \#Ent. & \#Rel. & \#Attr. & \#Rel tr. & \#Attr tr. & \#Image & \#Ref. \\
        \midrule
        \multirow{2}*{DBP15K$_{ZH-EN}$ {\scriptsize \cite{liu2021visual}}} & ZH & 19,388 & 1,701 & 8,111 & 70,414 & 248,035 & 15,912 & \multirow{2}*{15,000} \\
        & EN & 19,572 & 1,323 & 7,173 & 95,142 & 343,218 & 14,125 \\
        \midrule
        \multirow{2}*{DBP15K$_{JA-EN}$ {\scriptsize \cite{liu2021visual}}} & JA & 19,814 & 1,299 & 5,882 & 77,214 & 248,991 & 12,739 & \multirow{2}*{15,000} \\
        & EN & 19,780 & 1,153 & 6,066 & 93,484 & 320,616 & 13,741 \\
        \midrule
        \multirow{2}*{DBP15K$_{FR-EN}$ {\scriptsize \cite{liu2021visual}}} & FR & 19,661 & 903 & 4,547 & 105,998 & 273,825 & 14,174 & \multirow{2}*{15,000} \\
        & EN & 19,993 & 1,208 & 6,422 & 115,722 & 351,094 & 13,858 \\
        \midrule
        \multirow{2}*{FB15K-DB15K {\scriptsize \cite{liu2019mmkg}}} & FB15K & 14,951 & 1,345 & 116 & 592,213 & 29,395 & 13,444 & \multirow{2}*{12,846} \\
        & DB15K & 12,842 & 279 & 225 & 89,197 & 48,080 & 12,837 \\
        \midrule
        \multirow{2}*{FB15K-YAGO15K {\scriptsize \cite{liu2019mmkg}}} & FB15K & 14,951 & 1,345 & 116 & 592,213 & 29,395 & 13,444 & \multirow{2}*{11,199} \\
        & YAGO15K & 15,404 & 32 & 7 & 122,886 & 23,532 & 11,194 \\
        \bottomrule
    \end{tabular}
    \caption{Dataset Statistics.}
    \label{tab:dataset}
\end{table*}

% In the main paper, we define the overall loss of the proposed model MCLEA as follows:
% \begin{gather}
%     \mathcal{L} = \mathcal{L}^{\text{ICL}}_o + 
%     \textstyle\sum_{m\in\mathcal{M}} \alpha_m\mathcal{L}^{\text{ICL}}_m + 
%     \textstyle\sum_{m\in\mathcal{M}} \beta_m\mathcal{L}^{\text{IAL}}_m,
%     \label{eq:total-loss}
% \end{gather}
% where $\mathcal{M}=\{g,r,a,n,v\}$, $\mathcal{L}^{\text{ICL}}_o$ denotes the ICL (intra-modal contrastive loss) operated on joint embedding, $\alpha_m$ and $\beta_m$ are hyper-parameters that balance the importance of ICL and IAL (inter-modal alignment loss) for $m$-th modality, respectively.
% However, searching for optimal hyper-parameters manually are intractable and expensive.

In this section, we treat Eq.~(\ref{eq:loss}) as a multi-task loss function and combine multiple objectives using homoscedastic uncertainty~\cite{kendall2018multi}, allowing us to automatically learn the relative weights of each loss.

Firstly, the ICL can actually be regarded as a classification loss with negative log-likelihood, i.e., predicting whether two entities are equivalent. Here, we rewrite the loss function of ICL as follows (for simplicity, here we omit the modality index and the inner-graph negative samples, and only consider the unidirectional version):
\begin{gather}
\begin{aligned}
    \mathcal{L}^{\text{ICL}} 
    &= - \mathbb{E}_{i\in\mathcal{B}} 
    \log q(e^i_1,e^i_2) \\
    &= - \mathbb{E}_{i\in\mathcal{B}} 
    \log P\left(c=1|f^{\mathbf{W}}(e^i_1,e^i_2)\right),
\end{aligned}
\end{gather}
where $c=1$ means that the two input entities are equivalent, otherwise $c=0$; $f^{\mathbf{W}}(\cdot,\cdot)$ is the model output with parameter $\mathbf{W}$.
Following~\cite{kendall2018multi}, we adapt the negative log-likelihood to squash a scaled version of the model output with an uncertainty scalar $\sigma$ through a softmax function:
\begin{gather}
\begin{small}
\begin{aligned}
    &-\log P\left(c=1|f^{\mathbf{W}}(e^i_1,e^i_2), \sigma\right) \\
    &= -\log \text{Softmax} \left(\frac{1}{\sigma^2}f^{\mathbf{W}}(e^i_1,e^i_2)\right) \\
    &= -\frac{1}{\sigma^2}f^{\mathbf{W}}(e^i_1,e^i_2)
    + \log \sum_{j\neq i}\exp\left(
    \frac{1}{\sigma^2}f^{\mathbf{W}}(e^i_1,e^j_2)
    \right),
    \label{eq:log}
\end{aligned}
\end{small}
\end{gather}
where $e^j_2$ with $j\neq i$ is the cross-graph negative samples defined in the main paper.

Applying the same assumption in~\cite{kendall2018multi}:
\begin{gather}
\begin{aligned}
    &\frac{1}{\sigma}\sum_{j\neq i}\exp\left(
    \frac{1}{\sigma^2}f^{\mathbf{W}}(e^i_1,e^j_2)
    \right) \\
    &\approx
    \left(
    \sum_{c'}\exp\left(f^{\mathbf{W}}(e^i_1,e^j_2)\right)
    \right)^{\frac{1}{\sigma^2}},
\end{aligned}
\end{gather}
we can simplify Eq.~(\ref{eq:log}) to:
\begin{gather}
\begin{small}
\begin{aligned}
    &-\log P\left(c=1|f^{\mathbf{W}}(e^i_1,e^i_2), \sigma\right) \\
    % &= -\frac{1}{\sigma^2}f^{\mathbf{W}}(x_i,y_i)
    % + \log \sum_{j\neq i}\exp\left(
    % \frac{1}{\sigma^2}f^{\mathbf{W}}(x_i,y_j)
    % \right) \\
    &\approx -\frac{1}{\sigma^2}f^{\mathbf{W}}(e^i_1,e^i_2)
    + \frac{1}{\sigma^2}\log\sum_{j\neq i}\exp(f^{\mathbf{W}}(e^i_1,e^j_2)) + \log(\sigma) \\
    &= -\frac{1}{\sigma^2} \log P\left(c=1|f^{\mathbf{W}}(e^i_1,e^i_2)\right) + \log(\sigma),
    % &= \frac{1}{\sigma^2} \mathcal{L}^{\text{ICL}}(x_i) + \log(\sigma)
    \label{eq:approx}
\end{aligned}
\end{small}
\end{gather}
where $\sigma$ can be interpreted as the relative weight of the loss and automatically learned with stochastic gradient descent.

On the other hand, the IAL defines the KL divergence over the output distribution
between joint embedding and uni-modal embedding (we omit the modality index and only consider the unidirectional version for simplicity):
\begin{gather}
\begin{aligned}
    \mathcal{L}^{\text{IAL}} &=\mathbb{E}_{i\in\mathcal{B}}\ 
    \text{KL} (q'_o(e^i_1,e^i_2)\ ||\ q'(e^i_1,e^i_2)) \\
    &= \mathbb{E}_{i\in\mathcal{B}}\ 
    q'_o(e^i_1,e^i_2)\log\frac{q'_o(e^i_1,e^i_2)}{q'(e^i_1,e^i_2)} \\
    &= \mathbb{E}_{i\in\mathcal{B}}\ 
    [q'_o(e^i_1,e^i_2)\log q'_o(e^i_1,e^i_2) \\
    &\quad\quad - q'_o(e^i_1,e^i_2)\log q'(e^i_1,e^i_2)],
\end{aligned}
\end{gather}
where $q'_o(e^i_1,e^i_2)$ and $q'(e^i_1,e^i_2)$ represent the output predictions of joint embedding and the uni-modal embedding, respectively.
Since we only back-propagate through $q'(e^i_1,e^i_2)$ in Eq.~(\ref{eq:ial}), $\mathcal{L}^{\text{IAL}}$ is equivalent to calculating the cross-entropy loss over the two distributions:
\begin{gather}
\begin{aligned}
    \mathcal{L}^{\text{IAL}} = - q'_o(e^i_1,e^i_2)\log q'(e^i_1,e^i_2).
\end{aligned}
\end{gather}

Therefore, similar to ICL, we can automatically learn the relative weight of IAL for each modality through task-dependent uncertainty.
As mentioned above, the total loss in Eq.~(\ref{eq:loss}) can be rewritten as:
\begin{gather}
\begin{aligned}
    \mathcal{L} 
    = \mathcal{L}^{\text{ICL}}_o &+ 
    \textstyle\sum_{m\in\mathcal{M}} \left(\frac{1}{\alpha^2_m}\mathcal{L}^{\text{ICL}}_m +\log{\alpha_m}\right) \\
    \qquad&+ 
    \textstyle\sum_{m\in\mathcal{M}} \left(\frac{1}{\beta^2_m}\mathcal{L}^{\text{IAL}}_m +\log{\beta_m}\right),
\end{aligned}
\end{gather}
where $\alpha_m, \beta_m$ are learnable parameters. Large $\alpha_m$ ($\beta_m$) will decrease the contribution of $\mathcal{L}^{\text{ICL}}_m$ ($\mathcal{L}^{\text{IAL}}_m$) for the $m$-th modality, whereas small $\alpha_m$ ($\beta_m$) will increase its contribution.

\section{Dataset Statistics}

The detailed dataset statistics are listed in Table~\ref{tab:dataset}, including the number of entities (\#Ent.), relations (\#Rel.), attributes (\#Attr.), number of relation triples (\#Rel tr.) and attribute triples (\#Attr tr.), number of images (\#Image), and number of reference entity alignments (\#Ref.).
It is worth noting that not all entities have the associated images or the equivalent counterparts in the other KG.

\end{document}